\pdfoutput=1

\documentclass[11pt]{article}

\usepackage{ACL2023}

\usepackage{times}
\usepackage{latexsym}
\usepackage{booktabs}

\usepackage[T1]{fontenc}

\usepackage[utf8]{inputenc}

\usepackage{microtype}

\usepackage{framed}
\usepackage{caption}
\usepackage{subcaption}
\usepackage{inconsolata}
\usepackage{svg}
\usepackage{multirow}
\usepackage{multicol}
\usepackage{hyperref}
\usepackage{array}
\usepackage{soul}
\usepackage{xcolor,colortbl}

\newcolumntype{L}{>{\centering\arraybackslash}m{3cm}}
\usepackage{enumitem}
\newcommand{\dataset}{{\texttt{Researchy Questions}~}}

\title{Researchy Questions: A Dataset of Multi-Perspective, Decompositional Questions for LLM Web Agents}

\author{Corby Rosset \\
    Microsoft \\
    \And
    Ho-Lam Chung \\
    National Taiwan Univ. \\
    \And
    Guanghui Qin \\
    Johns Hopkins Univ. \\
    \And
    Ethan C. Chau \\
    Microsoft \\
    \AND
    Zhuo Feng \\
    Microsoft \\
    \And
    Ahmed Awadallah \\
    Microsoft \\
    \And
    Jennifer Neville \\
    Microsoft \\
    \And
    Nikhil Rao \\
    Microsoft \\
}

\begin{document}
\maketitle

\begin{abstract}
Existing question answering (QA) datasets are no longer challenging to most powerful Large Language Models (LLMs). 
Traditional QA benchmarks like TriviaQA, NaturalQuestions, ELI5 and HotpotQA mainly study ``known unknowns'' with clear indications of both what information is missing, and how to find it to answer the question. Hence, good performance on these benchmarks provides a false sense of security. 
A yet unmet need of the NLP community is a bank of non-factoid, multi-perspective questions involving a great deal of unclear information needs, i.e. ``unknown uknowns''. We claim we can find such questions in search engine logs, which is surprising because most question-intent queries are indeed factoid. 
We present \textbf{Researchy Questions}, a dataset of search engine queries tediously filtered to be non-factoid, ``decompositional'' and multi-perspective. 
We show that users spend a lot of ``effort'' on these questions in terms of signals like clicks and session length, and that they are also challenging for GPT-4. 
We also show that ``slow thinking'' answering techniques, like decomposition into sub-questions shows benefit over answering directly. 
We release\footnote{\url{https://huggingface.co/datasets/corbyrosset/researchy_questions}} $\sim$ 100k Researchy Questions, along with the Clueweb22 URLs that were clicked. 

\end{abstract}

\section{Introduction}


\begin{table}[ht]
\centering
\small
\setlength{\tabcolsep}{3pt} 
\begin{tabular}{c|cccc}
Dataset & Num. & Topics & Sub-Ques. & Sub-Query \\
 \hline
Hotpot QA  & 300 & 2.9 & 3.8 & 3.6            \\
OpenBook QA  & 300 & 3.8 & 6.3 & 5.9          \\
Strategy QA  & 300 & 3.8 & 5.3 & 4.9        \\
Truthful QA  & 300 & 3.8 & 6.4 & 6.0        \\
Aquamuse & 300 & 3.7 & 5.4 & 5.2           \\ \hline
Reddit/askh  & 300 & 4.9 & 9.4 & 8.5              \\
Reddit/asks & 300 & 5.1 & 9.2 & 8.8             \\
Reddit/eli5 & 300 & 4.5 & 9.7 & 9.3             \\
Stack Exchange & 300 & \textbf{6.1} & 8.4 & 7.6        \\
Wikihow     & 300 & 4.8 & 11.7 & 11.2         \\ \hline
\rowcolor{yellow} Researchy &  96k & 3.9 & \textbf{14.3} & \textbf{12.6}                
\\
\end{tabular}
\caption{We asked GPT-4 to decompose questions into a natural hierachy of subquestions, as well as concrete queries it would issue to a search engine. 
As expected, HotpotQA requires the fewest sub-questions to answer.}
\label{tab:closed_book_decompositions}
\end{table}

\begin{figure*}[h]
\centering
\includegraphics[width=\textwidth]{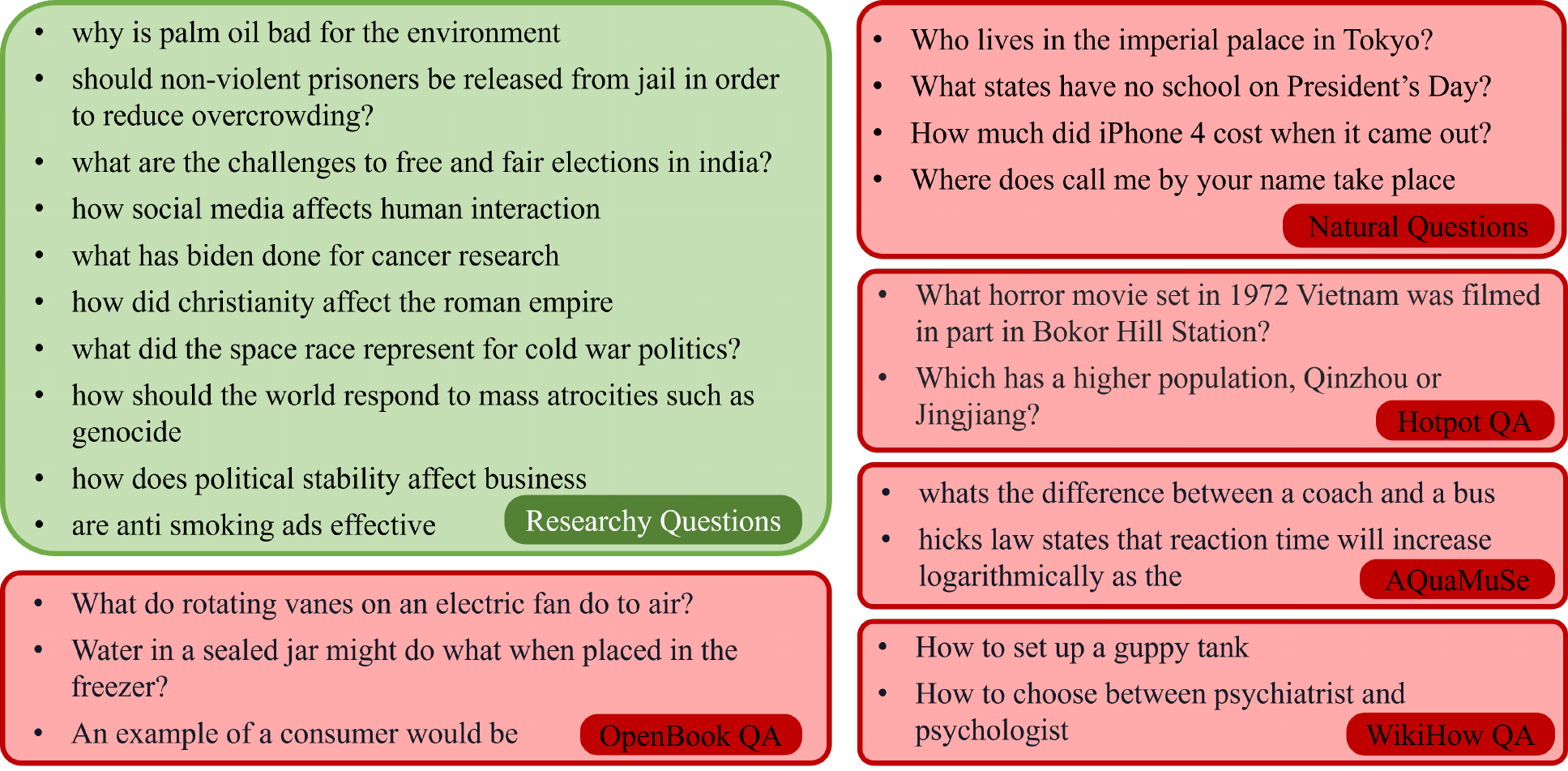}
\caption{Qualitative comparison of how \dataset differs from other Question Answering datasets. Researchy Questions involve a greater deal of complexity and ``unknown unknowns'' than other QA datasets. }
\label{fig:data-preview}
\end{figure*}

The advent of Large Language Models (LLMs) has ushered in a new era for the field natural language processing, with short- and long- form question-answering (QA) at the forefront of recent achievements~\cite{openai2023gpt4}. Historically, QA benchmarks served as the crucible for evaluating a model's grasp of natural language understanding. However, LLMs have nearly perfected many QA datasets, particularly those involving answering short, factual questions like \emph{``What is the capital of Brunei?"}. The re-emergence of chatbots and ``agentic'' AI assistants, which represent complex systems based on LLMs as a building block, have led to new opportunities for users to ask deeper, more nuanced questions as shown in Figure~\ref{fig:data-preview}.

However, the capabilities of powerful AI-assisted QA tools has outpaced the metrics needed to evaluate them. A number of traditional QA benchmarks like Natural Questions~\cite{natural_questions}, TriviaQA~\cite{2017arXivtriviaqa}, WebQuestions~\cite{berant-etal-2013-semantic}, SearchQA~\cite{dunn2017searchqa} have been more or less solved by modern LLMs. These datasets consist primarily of factoid questions (from search engine logs, Trivia, Jeopardy!, etc), where answers are typically found within a single sentence or paragraph that almost certainly exist in standard pre-training web corpora~\cite{zhou2023dont}. For instance, MS Marco ~\cite{Campos2016MSMA} is up to 55\% factoid questions~\citep{10.1145/3477495.3531926}. The obvious drawback to these kinds of QA datasets is that the answers can be memorized by LLMs or solved with simple pattern-matching or keyword-based search.

Multi-hop reasoning tasks including HotpotQA \cite{yang2018hotpotqa}, HybridQA~\cite{chen-etal-2020-hybridqa}, MuSiQue~\cite{trivedi2021musique} aim to challenge QA systems to logically bridge information across multiple documents or paragraphs.
While these datasets have made strides in increasing question complexity, the answers are still ultimately factoid, and it's clear what sub-questions ought to be asked to recall the missing information. Furthermore, the construction of these datasets (e.g. synthetically generated from paths of wikipedia links) leads to a distribution mismatch w.r.t questions humans ask.

There exist several sources of long-form, non-factoid QA datasets like ELI5~\cite{eli5_lfqa}, Stack Exchange, Yahoo Answers~\cite{zhang2016characterlevel}, and WikiHowQA~\cite{bolotova-baranova-etal-2023-wikihowqa}. While answers to these questions are more involved than factoid questions, ELI5 and WikiHowQA elicit more expository answers than analytical ones. The Oxford Allsouls dataset~\cite{liu2023evaluating} contains 1k college-level essay prompts which are multi-perspective, but are designed to evaluate persuasive writing skills and do not have associated documents to ground responses. AQuaMuSe~\cite{kulkarni2020aquamuse} is an excellent attempt at filtering Natural Questions (NQ) for multi-faceted queries, but their method was constrained by relying on the relatively short paragraph-length answers already present in NQ.


\begin{table*}[]
\small
\setlength{\tabcolsep}{4pt} 
\renewcommand{\arraystretch}{1.5} 
\begin{tabular}{|p{11.5cm}|p{4cm}|}
\hline
\multicolumn{2}{|l|}{\textbf{Researchy Question}: how does public transport help the economy}  \\ \hline
\multicolumn{1}{|c|}{\textbf{Hierarchical Decomposition of the Question}} & \multicolumn{1}{c|}{\textbf{Clicked Clueweb22 URLs}} \\ \hline
\begin{minipage}[t]{11.5cm}
{\begin{enumerate}[nolistsep,leftmargin=14pt]
    \item What is public transport?
    \begin{enumerate}[nolistsep,leftmargin=14pt]
        \item  What are the different types of public transport?
        \item How many people use public transport in different regions or countries?
    \end{enumerate}
    \item What are the direct economic benefits of public transport?
    \begin{enumerate}[nolistsep,leftmargin=14pt]
        \item How public transport reduces costs for users e.g. fuel, parking, maintenance, etc.?
        \item How does public transport generate revenue...
    \end{enumerate}
    \item What are the indirect economic benefits of public transport?
    \begin{enumerate}[nolistsep,leftmargin=14pt]
        \item How does public transport reduce congestion...
        \item How does public transport increase access to education, employment, health, ...
        \item How does public transport enhance productivity and innovation...
        \item How does public transport contribute to environmental and social goals...
    \end{enumerate}
    \item How do the economic benefits of public transport compare to the costs of providing and maintaining public transport?
    \begin{enumerate}[nolistsep,leftmargin=14pt]
        \item What are the main costs of public transport...
        \item How are the costs of public transport financed...
        \item How are the benefits and costs of public transport measured and evaluated...
    \end{enumerate}
\end{enumerate}
} 
\end{minipage} & \begin{minipage}[t]{4cm}
{
\begin{enumerate}[nolistsep,leftmargin=14pt]
    \item \href{https://infrastructureusa.org/the-economic-impact-of-public-transportation}{infrastructureusa.org}
    \item \href{https://nationalgeographic.org/article/effects-transportation-economy}{nationalgeographic.org}
    \item \href{https://quora.com/how-does-public-transportation-help-the-economy}{quora.com}
    \item \href{https://accessmagazine.org/spring-2012/can-public-transportation-increase-economic-efficiency}{accessmagazine.org}
    \item \href{https://frameworks.ced.berkeley.edu/2014/the-economic-benefits-of-transit-service}{ced.berkeley.edu}
    \item \href{https://greenertransportsolutions.com/guidance-tool/relationship-between-transport-economy}{greenertransportsolutions}
    \item \href{https://bts.gov/topics/transportation-and-economy}{bts.gov}
    \item \href{https://apta.com/research-technical-resources/economic-impact-of-public-transit}{apta.com}
\end{enumerate}
} 
\end{minipage} \\
\hline
\multicolumn{2}{|c|}
{\begin{minipage}[t]{15.5cm}
{Example of a \textbf{Pivotal Fact} from URL: \href{https://accessmagazine.org/spring-2012/can-public-transportation-increase-economic-efficiency}{accessmagazine.org} \\ \textit{``...Even in cities with a high concentration of office space in the [central business district], we estimate that \hl{increasing transit ridership by 10 percent will increase office rents by no more than 0.5 percent}. For all other cities, we estimate that increasing transit ridership will have no effect on office rents...''}} 
\end{minipage} 
} \\
\bottomrule
\end{tabular}
\caption{An example Researchy Question, how GPT-4 decomposed it into sub-questions (closed-book), as well the URLs in ClueWeb22 that real users clicked on, and an example of a \textbf{Pivotal Fact} from one of those URLs.
}
\label{tab:example_decomposition}
\end{table*}

The rise of ``LLM Agents'' e.g. ~\cite{wu2023autogen} have opened the door for even deeper collaboration between users, LLMs, and tools. In response, more recent datasets have focused on completing challenging tasks in open-ended environments with tools like a web browser, file system, database, etc. 

In particular, Gaia~\cite{mialon2023gaia} tests understanding of multi-modal inputs (images and text), as well as complex reasoning across levels of difficulty to solve. AgentBench~\cite{liu2023agentbench} provides closed environments for an LLM to interact with APIs in various scenarios, including coding (interacting with file systems or databases), games/puzzles, and web browsing/shopping.
While these datasets advance the field of metrics for LLM agents, they are small, consisting of only 466 and 1,091 questions respectively that were hand-curated by the authors.

The need for more challenging QA datasets also stems from some alarming trends: while there are hundreds of public LLMs, they are pretrained on only a handful existing corpora~\cite{gao2020pile, raffel2023exploring}, or distilled from one of a few teacher LLMs~\cite{peng2023instruction}. Furthermore, much more data scraped from the internet for training is itself going to be AI-generated content, leading to an echo-chamber effect~\cite{dohmatob2024tale, wu2023aigenerated}. \textbf{Hence, the convergent evolution~\cite{Stayton2015} of LLMs increases the risk that they will \textit{not} recognize that they \textit{don't} know something}, e.g. see GPT-4 and Mixtral 8x7b in Table~\ref{tab:closed-book-examples}. This is especially true when LLMs act in the capacity of LLM-as-a-judge~\cite{zheng2023judging, yuan2024selfrewarding}, or when faced with very multi-faceted/multi-perspective questions, where the consequences could be that users ``miss the whole picture'' or worse, are misled~\cite{zheng2023does, liu2023evaluating}. While retrieval-augmentation~\cite{lewis2021retrievalaugmented, borgeaud2022improving, guu2020realm} can help complement LLM agents, the risk is merely shifted to whether a sub-system retrieves the \textit{right} information and uses it correctly~\cite{liu2023lost}.  

We believe the well-studied phenomenon of ``\textbf{unknown unknowns}''~\cite{uscongress1981nasa} applies to LLM Agents in scenarios addressing complex questions requiring ``slow thinking''~\cite{Kahneman2011ThinkingFastSlow}. Simply put, one strategy is to iteratively re-frame or decompose the problem into a set of ``\textbf{known unknowns}'' (which characterize most of the aforementioned QA datasets). For these sub-problems, it should be clearer what information is missing, how to find it, and once found, how the ``known known'' contributes to the final answer. Several techniques such as chain-of-thought question decomposition~\cite{radhakrishnan2023question} and tree-of-thought~\cite{yao2023tree} prompting take a similar approach to plan long-horizon solutions to complex problems. However, those studies still operate over traditional QA benchmarks like HotpotQA, or over simple games like crossword puzzles. Hence, the right benchmark of questions for these advanced decomposition techniques still does not exist for open-domain web scenarios~\cite{krishna2021hurdles}. 

We present \dataset to study the dynamics of how LLM agents handle unclear information needs associated with very complex questions. 
We define a \textbf{Researchy Question} as a \emph{non-factoid} question that expects a \emph{long-form answer} (longer than a paragraph!) entailing substantial research or effort to synthesize. A Researchy Question can be instantiated as a complex search task~\cite{complexandexploratory} with unclear information needs that requires analyzing \emph{multiple documents} or pieces of evidence. A Researchy Question does not have a single correct answer, but rather \emph{multiple perspectives} allowing a dense manifold of answers over which varying criteria can determine which is better. In practice, the act of answering a Researchy Question probably involves \emph{decomposition} into sub-questions that aid the retrieval of comprehensive information, reducing the risk of missing unknown unknowns.
Lastly, a Researchy Question represents a \emph{genuine information need} that real people asked. Figure~\ref{fig:data-preview} qualitatively compares other canonical QA datasets. 

\dataset is primarily a QA dataset to evaluate question answering systems or LLM Agents with the ultimate goal of achieving ever higher-quality answers using any tools necessary. Notwithstanding, it is also a search/retrieval dataset in the sense that finding and correctly incorporating the right evidence is a critical sub-system to meet expectations of trustworthiness and groundedness~\cite{zheng2023does, liu2023evaluating}. While we believe that question decomposition is a critical piece of solving \dataset, it is unclear how to define or measure the quality of sub-questions. To aid this endeavor, we reveal what URLs end users found useful, with the hope that good sub-questions will at least lead to the information found in those clicked documents.  

We release about 96K \dataset consisting of real users' queries to a commercial search engine, and additionally:
\begin{enumerate}[nolistsep]
    \item The decomposition of the question into a 2-level hierarchical plan (See Table~\ref{tab:example_decomposition} Left).
    \item For each question, the user-aggregated click distribution over URLs in a publicly available web corpus, ClueWeb22.
    \item Ordered list of sub-queries corresponding roughly to the sub-questions that can be readily issued to a search engine
\end{enumerate}

 In Section~\ref{sec:Construction} we describe how \dataset were obtained and then characterize them in Section~\ref{sec:CharacterizingResearchyQuestions}. In Section~\ref{sec:AgreementWithSearchBehavior} we verify that web users expend more effort on \dataset over other queries. In Section~\ref{sec:evaluation} we evaluate and compare decompositional answering techniques from~\cite{radhakrishnan2023question}.

\section{Researchy Questions Construction}
\label{sec:Construction}

\begin{table}[]
\small
\setlength{\tabcolsep}{3pt} 
\renewcommand{\arraystretch}{1.0} 
\begin{tabular}{r|l|ccc}
First             & MS Marco     & QnA                     & Non-factoid & Researchy ($\downarrow$) \\ \hline
how                    & 17.0\%        & 34.3\%                  & 29.4\%      & 41.2\%                   \\
why                    & \textcolor{red}{1.64\%}        & 6.26\%                  & 33.4\%      & 22.9\%                   \\
what                   & 34.9\%        & 12.2\%                  & 21.3\%      & 19.1\%                   \\
is/are/do              & 5.77\%        & 15.0\%                 & 6.50\%      & 9.67\%                   \\
should                 & 0.11\%        & 0.53\%                  & 0.59\%      & 1.85\%                   \\
can                    & 1.84\%        & 4.31\%                  & 1.01\%      & 0.97\%                   \\
\textcolor{red}{who}   & 3.27\%        & 4.77\% & 0.90\%      & 0.47\%                   \\
which                  & 1.78\%        & 2.78\%                  & 1.87\%      & 0.46\%                   \\
\textcolor{red}{when}  & 2.70\%        & 5.03\% & 0.44\%      & 0.43\%                   \\
pros                   & $<$ 0.01\% & 0.03\%                  & 0.45\%      & 0.28\%                   \\
explain                & 0.05\%        & 0.06\%                  & 0.12\%      & 0.23\%                   \\
\textcolor{red}{where} & 3.54\%        & 4.08\% & 0.37\%      & 0.17\%                   \\
factors                & $<$ 0.01\% & 0.01\%                  & 0.08\%      & 0.15\%                   \\
will                   & 0.10\%        & 0.69\%                  & 0.08\%      & 0.15\%                   \\
describe               & 0.05\%        & 0.07\%                  & 0.04\%      & 0.09\%                  
\end{tabular}
\caption{Top most-prevalent first-words of the three main stages of our question filtering funnel (with MS Marco for comparison). 
}
\label{tab:distribution_first_words}
\end{table}

\dataset are real user queries from search logs. While search logs have a rich diversity of query types and intents~\cite{10.1145/3477495.3531926, bu-etal-2010-function}, they mostly contain factoid or navigational queries, which need to be filtered. 

\subsection{Stage 1: Mining Search Logs}
\begin{table*}[t]
\renewcommand{\arraystretch}{1.1} 
\small
\setlength{\tabcolsep}{2pt} 
\begin{minipage}{.5\linewidth}
\centering
\begin{tabular}{c|ccccc}
        \textbf{Query Type} & Count &\begin{tabular}{@{}c@{}}Avg. unique \\ clicked urls\end{tabular}  & \begin{tabular}{@{}c@{}} Avg. \\ clicks \end{tabular} & \begin{tabular}{@{}c@{}} Avg. sat. \\ clicks \end{tabular} \\\hline
        General Queries & $\geq 1B$ & 1.88 & 4.83 & 2.54  \\ 
        QnA Queries & 15.7M & 3.99 & 9.31 & 5.10  \\ 
        Non-factoid QnA & 1.0M & 4.20 & 8.99 & 4.86  \\ 
        Researchy Queries & 100k & \textbf{6.31} & \textbf{15.85} & \textbf{8.54} \\
        \hline
\end{tabular}
\end{minipage}%
\begin{minipage}{.5\linewidth}
\centering
\begin{tabular}{c|cccc}
        \textbf{Session Type} & Count & \# Turns  & \begin{tabular}{@{}c@{}} \# unique. \\ queries \end{tabular} & \begin{tabular}{@{}c@{}} \# sat \\ clicks \end{tabular}  \\\hline
        General Sessions & $\geq 10B$
        & 2.42 & 2.11 & 0.76 \\ 
        QnA Sessions & $\geq 100M$
        & 6.28 & 5.53 & 1.15  \\ 
        N.F. QnA Sessions & $\geq 10M$
        & 12.89 & 11.33 & 1.91 \\ 
        Researchy Sessions & $\geq 1M$
        & \textbf{13.45} & \textbf{11.81} & \textbf{2.46} \\
        \hline
    \end{tabular}
\end{minipage}%
\caption{Our query filtering funnel; each row is a subset from the row above it. (Left) \dataset have much more clicks (Sat clicks have longer dwelltime), and require more unique documents;  the full distribution is in Figure~\ref{fig:queries_docs_histogram}. (Right) Search statistics \textit{per session in which each type of query appeared}, indicating that harder questions appear in longer sessions. These behaviors confirm our filtering yields more complex questions.}
\label{tab:user_log_analyze}
\end{table*}



We obtained a set of query-url click pairs from a commercial search engine that were logged between July 2021 and August 2022, which maximally overlaps with the creation of the Clueweb22 snapshot of web documents~\cite{overwijk2022clueweb22}. This way, we can simply indicate which \dataset clicked on which documents. We began with a large sample of english, non-adult queries that had least one click. We denote these as ``\textbf{General Queries}'', which were further filtered.

An important filtering criteria is frequency: we kept queries if they were issued at least 50 times in the logs. This criteria is simple but powerful: it helps denoise the dataset (reduces spelling errors), and also allows us to focus on questions that are not ``one-offs''. This helps us glean insights into repeated user behaviors as they interact with the search engine.

In order to select queries which are answer-seeking intent (i.e. actual ``questions'' as distinct from navigational queries like ``\textit{facebook login}'', shopping intent like ``\textit{fast running shoes}'', or local intent like ``\textit{best pizza near me}''), we used a suite of rules and existing production classifiers:
\begin{itemize}[nolistsep]
    \item Query Language: English
    \item Adult intent: False
    \item Number of Distinct Occurrences: $\geq$ 50
    \item 3 $\leq$ Number of Query Words: $\leq$ 15
    \item Number of Distinct URLs Clicked On: $\geq 2$
    \item Question Intent Classifier: True
    \item Navigational Intent: False
    \item Local / Real Estate / Map Intent: False
    \item Retail/Shopping Intent: False
    \item Coding/Technical Intent: False
    \item Health/Medical Intent: False
    \item Triggered Possible Answer Cards: $\geq 1$
    \item Triggered lots of Ads: False
\end{itemize}

To explain some of the points above: an Answer Card is a high-precision feature in a search engine where a paragraph containing the answer is shown at the top of the results page, distinct from the ``ten blue links''. 
Because search engines are constantly being updated with new features and triggering rules, all the above statistics are aggregated and normalized over the full year period. For instance, ``a lot of Ads'' is determined by summing the total number of ads shown for the query over the full year and dividing by the number of times the query was issued, and then picking a threshold above which queries seemed ``shopping intent''. The Ads requirement also helped catch any shopping intent queries missed by the Retail intent classifier. 

We wanted to remove Coding/Technical queries because such questions are often motivated by very specific problems that are often solved by one piece of documentation after lots of clicking around, which is not the behavior we wish to target in this dataset. Health and medical questions are avoided largely because they often overlap too much with those that ought to be addressed by a licensed medical professional. Many shopping / retail queries could be construed as ``researchy'' e.g. ``what are the best headphones'', but we avoid them in this dataset because it is hard to distinguish whether a URL was clicked due to aggressive advertising or a real information need. 

After this stage of filtering, we arrived at 15.7M ``\textbf{QnA Queries}'' which are largely recognizable as natural language questions that deal with open-domain knowledge. This size was manageable enough to efficiently run our own bert-large scale classifiers we use in the next stage of filtering.  

\subsection{Stage 2: Factoid Classifier}
We needed a way to distinguish which QnA Queries were factoid vs. non-factoid; we trained a binary classifier on auto-labeled data for this purpose. The training data was a uniform sample of 200k questions from the 15.7M QnA Queries. Labels for the questions were collected from gpt3 (\texttt{text-davinci-003}) prompted with few-shot examples as shown in Figure~\ref{fig:promp-label-nonfactoid}. The labels were then used to train a \texttt{bert-large} non-factoid question classifier, which was then inferenced on the full set of 15.7M queries. By manual inspection, a threshold was chosen above which we were satisfied that the questions were meaningfully non-factoid. The resulting 1.0M met the non-factoid threshold of 0.75 as shown on the left hand side of Figure~\ref{fig:histograms_factoid_decomposition}, which we denote as ''\textbf{Non-factoid QnA Queries}''. 

\subsection{Stage 3: Decompositional Classifier}
Not all the resulting non-factoid QnA queries exhibited the ``decompositional'' information needs. Namely, they often looked expository or ``how-to'' with generally one correct answer which did not have many perspectives. We trained a second classifier to score the extent to which a question requires asking sub-questions. The specific definition of ``requiring sub-questions'' is described in a prompt in Figure~\ref{fig:promp-label-decompositional} given to ChatGPT (\texttt{gpt-35-turbo}) to collect labels. We used ChatGPT because we figured this was a relatively intense cognitive task. We inferenced ChatGPT on about 40k outputs from the non-factoid classifier that met the non-factoid threshold of 0.75. We used the labels to train a separate \texttt{bert-large} ``decompositional'' classifier. 

Again, we chose a threshold by manual inspection to indicate which of the 1.0M \textbf{Non-factoid QnA Queries} were also decompositional, which happened to be 0.6 as shown on the right-hand side of Figure~\ref{fig:histograms_factoid_decomposition}. Out of the 1.0M queries that met the non-factoid threshold, 146k also met the decompositional threshold. These 146k became the \dataset candidates \textit{before} deduplication. This procedure is not perfect; in Table~\ref{tab:nonfactoid_but_not_quite_decompositional} we show examples of some non-factoid questions that did not meet the decompositional threshold. 

\subsection{Stage 4: Deduplication}
The final step of filtering is deduplication.
We take an agglomerative clustering approach ~\cite{clusteranalysis}, where the only parameter is a distance threshold $\epsilon$ below which two queries are considered ``duplicate intent''. 

We represent the semantic intent of a query by an ANCE-based~\cite{xiong2020approximate} vector encoder $\overrightarrow{q_i} \leftarrow \texttt{encoder}(q_i)$. We instantiate a metric space defined by $1 - \texttt{cosine}(\cdot, \cdot)$ of the vector encodings using the faiss implementation of an approximate nearest neighbor (ANN) index~\cite{johnson2019billion}. For each question in the index, we search for the nearest neighbors $\{q_j \sim \texttt{ANN}(q_i)$ s.t. 1.0 - $\overrightarrow{q_i} \cdot \overrightarrow{q_j} < \epsilon\}$. For agglomerative clustering, we define a ``group'' as a set of queries in which all pairwise distances are within $\epsilon$. We found that about 63\% of queries were singletons (did not belong to a group of size greater than one), and the average group size was 3.8. For example, the queries ``\emph{what were tanks used for in ww1}'', ``\emph{how were the tanks used in ww1}'' and ``\emph{why were tanks needed in ww1}'' were all part of the same group. For all groups of size greater than one, we selected the query that was issued most often in the logs as the representative ``head'' of the group. After combining the heads of groups and singletons, about 70\% of queries remained, yielding 102k \dataset. Although we did our best to de-duplicate question intent, some clusters of topics remain, e.g. a quick keyword count shows about 600 contain the string ``ww2'', or about 80 for ``supreme court''.  

\subsection{Stage 5: Final GPT-4 Filtering}
\label{sec:stage-5-filtering}
As a final quality control step after deduplication, we had all 102k questions labeled by GPT-4 for intrinsic attributes of the question like how \textbf{multi-faceted} it is, how \textbf{reasoning-intensive} it is likely to be, etc. The full set of eight attributes is defined in Figure~\ref{fig:gpt4-classifier-prompt}, along with histograms of those scores in Figure~\ref{fig:autolabeling_histograms_gpt} for both \dataset and Natural Questions. All eight are scored on a scale of 1-10. About 3\% of 102k questions were removed based on the attributes ``\textbf{ambiguous}'' and ``\textbf{incomplete}'' which indicate defective questions too difficult to answer; some examples are shown in Table~\ref{tab:example_incomplete_queries}. Another 2\% were removed for being too ``\textbf{assumptive}'', meaning the question was phrased in a presumptuous way that was likely to bias the answer as shown in Table ~\ref{tab:example_assumptive_queries}. 
Another 2\% was removed for safety reasons as shown in Table~\ref{tab:example_harmful_queries}, where we deem risk of harm too high by attempting to answer the question. Not all ``assumptive'' questions are harmful per se. Finally, the remaining 96k queries are what we release. 

\begin{table*}[]
\centering
\small
\setlength{\tabcolsep}{6pt} 
\renewcommand{\arraystretch}{1.1} 
\begin{tabular}{c|ccccccc|c}
\multirow{2}{*}{Method} &\multirow{2}{*}{Samples} & \multicolumn{2}{c}{Direct Answer} & \multicolumn{2}{c}{CoT decomp.} & \multicolumn{2}{c}{Factor decomp.} & Decomp.  \\
&    & Acc.        & Score             & Acc.         & Score              & Acc.          & Score                & Score gain\\ \hline
Hotpot QA               & 300                   & 0.843      & 83.4              & \textbf{0.877}       & 83.5      & 0.837        & 81.3                 & +0.1               \\
OpenBook QA            & 300                   & \textbf{0.926}      & 86.1   & 0.843       & 83.5               & 0.750         & 80.7                 & -2.6               \\
Strategy QA          & 300                   & 0.757      & 80.8              & \textbf{0.810}        & 83.7      & 0.777        & 82.6                 & +2.9               \\
Truthful QA          & 300                   & 0.703      & 73.7              & \textbf{0.789}       & 82.4      & 0.739        & 81.5                 & \textbf{+8.7}               \\
Aquamuse             & 300                   & 0.916      & 83.0              & \textbf{0.940}        & 84.9               & 0.926        & 85.0        & +2.0               \\ \hline
Reddit/askh               & 300                   & 0.759      & \textbf{79.8}              & 0.736       & 77.3               & 0.732        & 79.3                 & -0.5               \\
Reddit/asks               & 300                   & 0.783      & 81.1              & 0.743       & 79.4               & 0.796        & \textbf{82.7}                 & +1.6               \\
Reddit/eli5               & 300                   & 0.883      & 83.1              & 0.890        & 85.0               & 0.890         & \textbf{86.5}                 & +3.4               \\
StackExchange        & 300                   & 0.717      & \textbf{78.4}              & 0.599       & 70.7               & 0.628        & 73.4                 & -5.0               \\
Wikihow QA             & 300                   & 0.93       & 82.9              & 0.937       & 84.4               & 0.950         & \textbf{88.2}                 & +5.3               \\ 
Researchy Questions     & 1k                    & N/A        & 82.7              & N/A         & 84.6               & N/A          & \textbf{88.3}                 & +\textbf{5.6}              
\\ 
\end{tabular}
\caption{Comparison of various question decomposition techniques to prompt GPT-4 as an answer module. The top half of the table is short-form questions, the bottom half is longform questions. 
}
\label{tab:eval_decomposition_techniques}
\end{table*}

\section{Characterizing Researchy Questions}
\label{sec:CharacterizingResearchyQuestions}
Across the 96k \dataset (split into 90k train, 6.4k test), there are in aggregate 350k unique documents clicked, 48\% of which can be found in the english subset of Clueweb22 Set B~\cite{overwijk2022clueweb22}; the rest are in Sets A or L. For each question we release, there are on average 4.9 +/- 3.5 clicked documents (See Figure~\ref{fig:queries_docs_histogram} Right), indicating a good diversity of information needs and much higher than the average query overall. Conversely, for each document there is only 1.4 +/- 2.3 associated \dataset (See Figure~\ref{fig:queries_docs_histogram} Left) indicating good query de-duplication. 

To get a sense of how intrinsically difficult \dataset are compared to other datasets, we asked GPT-4 how many sub-questions or search engine queries would be necessary to ask/issue in order to fully answer each question. An example decomposition in shown in Table~\ref{tab:example_decomposition}, and the aggregate results are shown in Table~\ref{tab:closed_book_decompositions}. Clearly, GPT-4 thinks that most factoid QA datasets (top of the table) require the fewest sub-questions to answer, whereas \dataset require the most even among longer-form QA datasets. 

We also compared \dataset to another search-log based QA dataset -- Natural Questions~\cite{natural_questions} -- along the 8 quality dimensions descried in Section~\ref{sec:stage-5-filtering}, such as how reasoning- and knowledge-intense they are. The comparative histograms are shown in Figure~\ref{fig:autolabeling_histograms_gpt}, Clearly, GPT-4 thinks that \dataset require much more knowledge, reasoning, and are inherently more multi-faceted. 

Table~\ref{tab:distribution_first_words} shows the distribution of first words across \dataset (as well as intermediate datasets used in the filtering funnel described in the next section). For comparison, MS Marco queries (also from web search logs) are much more factoid -- e.g. only 1.64\% of them start with ``why''~\cite{bajaj2018ms}. 

Lastly, an emergent property of \dataset we observed is that some information found in the clicked URLs is very surprising, e.g. for the question ``\textit{should the death penalty be legalized}'', the fact that in the US, \textit{``the cost of enforcing the death penalty cost millions more than life imprisonment''} \footnote{\hyperlink{https://www.supremecourt.gov/opinions/urls_cited/ot2016/16-5247/16-5247-2.pdf}{supremecourt.gov/opinions}} would have a large impact on the economic arguments in that answer. We define a \textbf{Pivotal Fact} as a piece of information that is so surprising and consequential that, once known, drastically changes how an LLM Agent would answer the question (and the answer wouldn't be as good without it); but it wouldn't know about it unless it asked the right sub-question to retrieve it, e.g. ``\textit{does the death penalty cost more than life imprisonment}''. Hence, a pivotal fact is the information analogue of a black swan, a critical event that is hard to predict but highly impactful~\cite{Taleb2008BlackSwan}; another example is shown in the bottom of Table~\ref{tab:example_decomposition}. Although we don't have a good way to quantify the prevalence of \textbf{Pivotal Facts}, we believe that \dataset is the best dataset to study the dynamics of how LLM Agents search for, and respond to, these uknown unknowns.

\section{Agreement with User Search Behavior}
\label{sec:AgreementWithSearchBehavior}
More complicated questions ought to require more effort to answer~\cite{10.1145/2808194.2809465}. 
We can approximate the amount of effort users expend in terms of behavior signals like clicks and session length.

In Table~\ref{tab:user_log_analyze} (Left) we show aggregate click statistics for each subset of queries. The results show that \dataset, which are both non-factoid and decompositional, lead to more in-depth consumption (clicks and sat-clicks) of more diverse information (unique urls), agreeing with prior work~\cite{10.1145/2556195.2556221}. 

In Table~\ref{tab:user_log_analyze} (Right) we show behavioral signals exhibited by users at the session level rather than individual click level. For instance, if a QnA-type query appeared at any time in any session in the date range, the whole session is included in the ``QnA Sessions'' row. The results clearly show users were twice as engaged answering non-factoid questions than sessions devoted to factoid-y intents, and six times longer than the average session overall. 

\section{Evaluating Answer Techniques to Researchy Questions}
\label{sec:evaluation}

Since there is no one ``correct'' answer to a Researchy Question, we contend that they ought be evaluated in a relative, side-by-side fashion \emph{a la} Alpaca-Eval syle~\cite{alpaca_eval}, with e.g. the ``closed-book'' answer as the reference. 

Since \dataset are intended to be answered by decomposing them into sub-questions, we evaluate two decompositional question answering answer techniques -- chain-of-thought decomposition, and factored decomposition -- against the direct answering baseline. Factored decomposition makes a separate call to an LLM for each sub-question independently, and then a final ``recomposition'' call to synthesize the main answer~\citep{radhakrishnan2023question}. 

Table~\ref{tab:eval_decomposition_techniques} shows the results of a side-by-side auto-eval of the three answering techniques across a range of datasets. The LLM providing the answers was GPT-4, and the LLM-as-a-judge was also GPT-4, prompted separately using the template in Figure~\ref{fig:critique-then-score-prompt}. Since the gold correct answers exist for many of the datasets we consider, the judge was asked to ascertain ``Accuracy'' as a binary score of whether the candidate answer was consistent with the gold. The ``Score'' is on a scale of 1-100 indicating the overall quality.
The top half of Table~\ref{tab:eval_decomposition_techniques} corresponds to short-answer datasets (where Accuracy is more important), while the bottom half is longer answer questions where the overall Score is better suited to evaluate. Out of the long-form answer datasets, \dataset benefits the most from decomposition techniques. 

We draw several conclusions from Table~\ref{tab:eval_decomposition_techniques}. Firstly, we largely confirm the results of ~\cite{radhakrishnan2023question} that decompositional techniques improve the accuracy on short-form multi-hop datasets over zero-shot direct answering. 
Secondly, factored decomposition leads to the highest scores for long-form questions, particularly questions involving reasoning about complex processes like Wikihow and \dataset. On the other hand, chain-of-thought decomposition may be better suited to deduce correct answers to more logic-intensive questions. 
We believe the results for \dataset using factored decomposition would be much higher if we incorporated retrieved information.

\section{Related Work}
\subsection{Human Behavior in Search Sessions}
There has been some fundamental research into understanding user behavior in search sessions, from user studies~\cite{10.1145/2808194.2809465} to large scale click log evaluations~\cite{10.1145/2556195.2556221}. The latter attempt to ascertain signals that could indicate whether a user is ``struggling'' or ``exploring'' in their search session; we use many of the same signals. Namely, they conclude that ``exploring'' sessions contain more clicks because users want to locate information for multiple facets of a topic. We concur with their results, e.g. non-factoid questions involve more clicks than factoid ones. Similarly, the definition of a ``complex search task'' is almost congruent to that of a Researchy Question in its information seeking behaviors~\citeauthor{complexandexploratory}. Other studies try to identify complex search tasks and offer recommended sub-tasks~\cite{10.1145/2661829.2661912, zhang-etal-2021-learning}. Much of the intuition on how to filter \dataset stems from these and similar studies on user search behavior. 

\subsection{Iterative Retrieval-Augmented Generation}
Many papers discuss variants of chain-of-thought~\cite{wei2023chainofthought} adapted to decomposing multi-faceted questions such as query refinement prompts~\cite{amplayo2022query}, Decomposed Prompting~\cite{khot2023decomposed} and ReAct~\cite{yao2023react}. A step further is to marry a generative LLM with a vector-based retrieval system, e.g.~\cite{ren2023rocketqav2, xiong2020approximate, karpukhin2020dense, izacard-grave-2021-leveraging}. Such methods are instantiated in many flavors: Self-Ask~\cite{press2023measuring} was one of the first chain-of-thought-like decomposed prompting techniques to integrate a search engine, forcing an LLM to iteratively ask and subquestions. IRCoT interleaves retrieval with chain-of-thought such that what to retrieve next depends on what was previously retrieved~\cite{trivedi2023interleaving} Iter-RetGen iteratively retrieves and generates candidate answers to use as inputs to the next phase of retrieval~\cite{shao2023enhancing}, while Beam Retrieval maintains running hypotheses of the relevant passages at every step~\cite{zhang2023beam}. 

\subsection{Agentic Question Answering}
There are several ``agentic'' frameworks to facilitate the dynamic interaction between tools e.g. retrieval systems and LLMs-as-agents which are adept at iterative retrieval-augmented QA tasks. One is Demonstrate-Search-Predict~\cite{khattab2023demonstratesearchpredict}, another is AutoGen~\cite{wu2023autogen}. Some existing Web-based Agents include WebGPT~\cite{nakano2022webgpt} which imitate how users surf the web and ask queries, others like WebAgent~\cite{gur2023realworld} interact with the web programmatically by understanding the raw HTML. 

There are also several consumer-facing products for agentic search assistants such as Bing Chat~\footnote{\url{https://bing.com/chat}}, YouPro~\footnote{\url{https://you.com/search}} (Research Mode) and SciPhi~\footnote{\url{https://search.sciphi.ai/research}}. All of these systems clearly decompose queries into sub-questions and then retrieve/crawl the necessary pages to synthesize the final outputs.  An example of each addressing a Researchy Question is shown in Figures~\ref{fig:bingchat_example}, ~\ref{fig:youpro_research_example}, and ~\ref{fig:sciphi_research_example}, respectively.

\section{Conclusion}
For some time now, many QA datasets have been saturated, and many researchers have dismissed search logs as sources of complex questions. We present \dataset, a large dataset to advance the field of complex question answering over multiple documents and perspectives, targeting LLM-assisted web search agents. We detail how these complex queries are mined from search logs, and confirm they involve significantly more effort than other types of search queries. We also provide some preliminary evidence that decompositional answering techniques yield stronger results on \dataset than direct answering.

By design, these questions don't have a gold answer, so regrettably, it difficult to quantify the ``headroom'' of existing models, but qualitatively (e.g. Table~\ref{tab:closed-book-examples}) there seems to be substantial room for improvement. There also remains much work surrounding how to measure the quality of sub-questions, and studying how LLM agents find and interact with Pivotal Facts. We hope this dataset can help aid the development of new metrics and unlock new experiences for web users.

\clearpage
\section*{Limitations}
One of the primary limitations of this study is that while we argue question decomposition is key to solving complex \dataset, we don't propose a good way to measure the quality of a candidate set of sub-questions. Instead, we loosely argue that good sub-questions ought to at least lead to the same set of information found in the documents users clicked on for the question. We admit these two things are not the same, but clicks are a powerful signal. We try to make up for this gap by releasing the hierarchical question/query decomposition that GPT-4 gives closed-book, but we do not evaluate whether that decomposition does indeed lead to the same set of gold documents in a real retrieval system e.g. built from the entirety of ClueWeb22. 

We also admit that it would potentially be better to produce the question decomposition by ``working backwards'' from the documents that were clicked on -- i.e. identify key information in the clicked document for a researchy question and then identify what sub-question would lead to that information being retrieved. On the other hand, one could argue that the ``forward'' direction is what needs to be implemented in practice. While we do not address these points in this study, others in the academic community can explore these topics more with the dataset we release.  

Another limitation of this study is that Pivotal Facts are purely an observational phenomenon that we have in no way quantified. Future researchers could create a LLM-prompt to count the number of such statements in the clicked documents. 

We regret that this dataset is not multi-lingual. This is due to the large amount of uncertainty and trial-and-error in curating this dataset, which meant frequent manual inspection of data. We believe the same framework can be used to construct a multi-lingual version of \dataset. 

\section*{Ethics Statement}
We obtained approval to release this dataset after thorough IRB proceedings to ensure compliance with privacy, safety, and legal guidelines. 

We would like to make a few notes: while it may appear that we tried to remove queries that seemed ``controversial'', it is not our intention to act as moral or political regulators to decide if user queries are well-meaning. At the scale of web search, one will notice that users ask a lot of questions for a lot of reasons, and it is beyond the scope of our job to speculate what the motives are behind a query. It \emph{is} is our job to assess whether the act of attempting to answer a question will lead to a reasonable risk of harm. Also, it is not the case that ``GPT-4 is our moral authority'' on the topic of safety, it is merely one of many tools we employ in a stack to ensure our internal requirements are being met. 


\clearpage
\bibliography{custom}
\bibliographystyle{acl_natbib}

\appendix
\onecolumn
\section{GPT-4-based Filtering Details}

\begin{figure*}
    \centering
    \includegraphics[width=\linewidth]{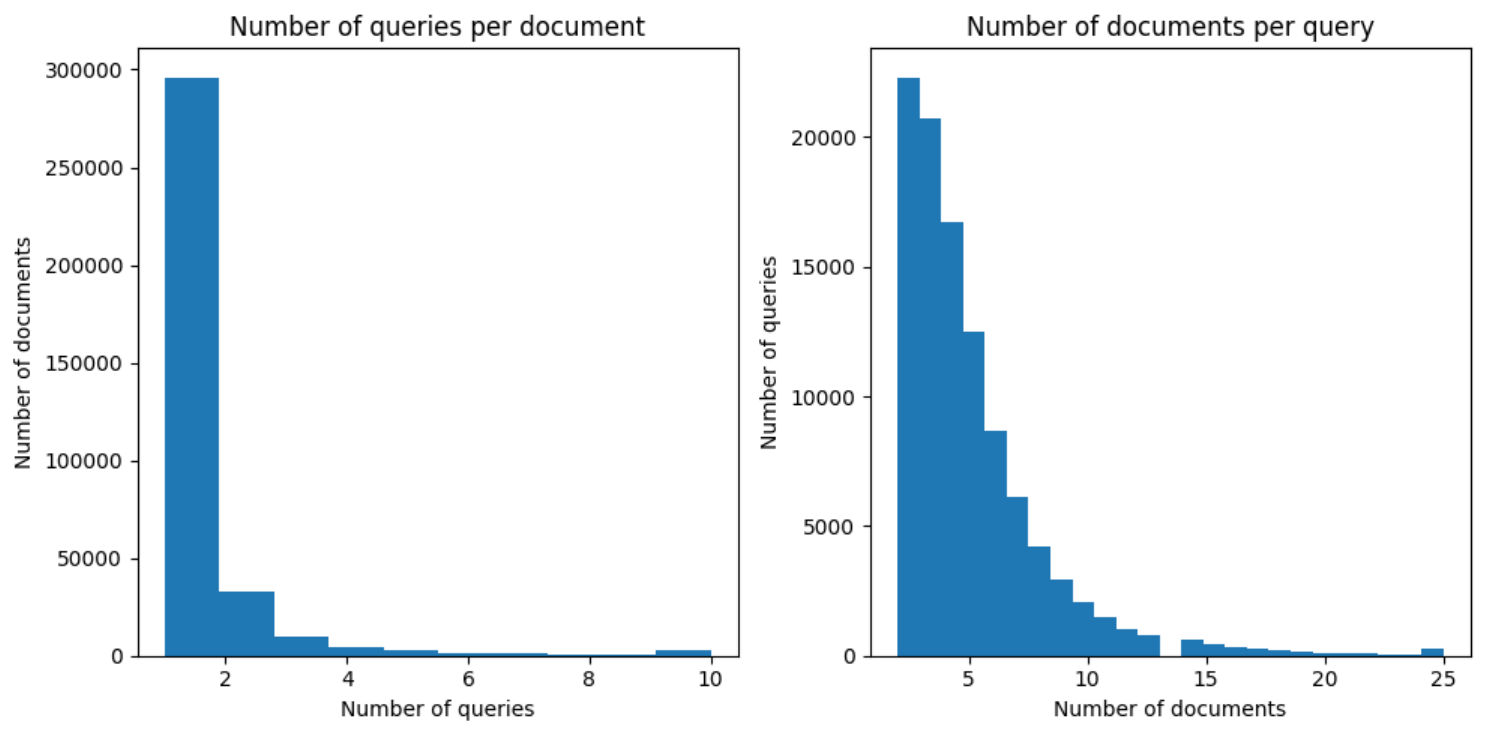}
    \caption{(Right) Histogram of number of documents clicked per question for \dataset which is much higher than for general web search queries. (Left) number of queries associated with each document. The fact that not very many queries are associated with each document validates the effectiveness of our query-deduplication procedure. }
    \label{fig:queries_docs_histogram}
\end{figure*}

\begin{figure*}
\centering
\begin{subfigure}{.5\textwidth}
  \centering
  \includegraphics[width=\linewidth]{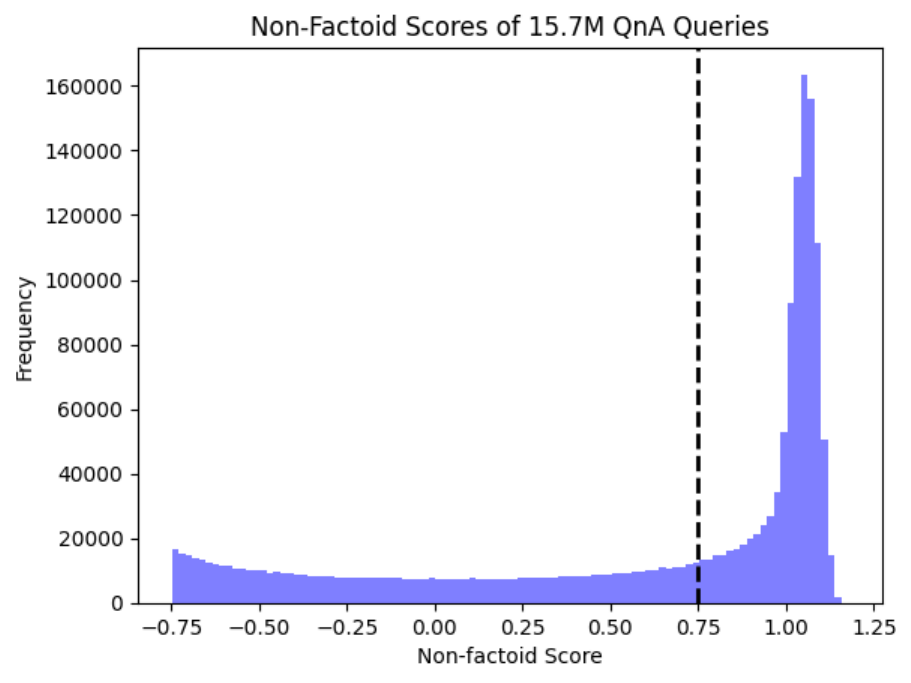}
  \label{fig:sub1}
\end{subfigure}%
\begin{subfigure}{.5\textwidth}
  \centering
  \includegraphics[width=\linewidth]{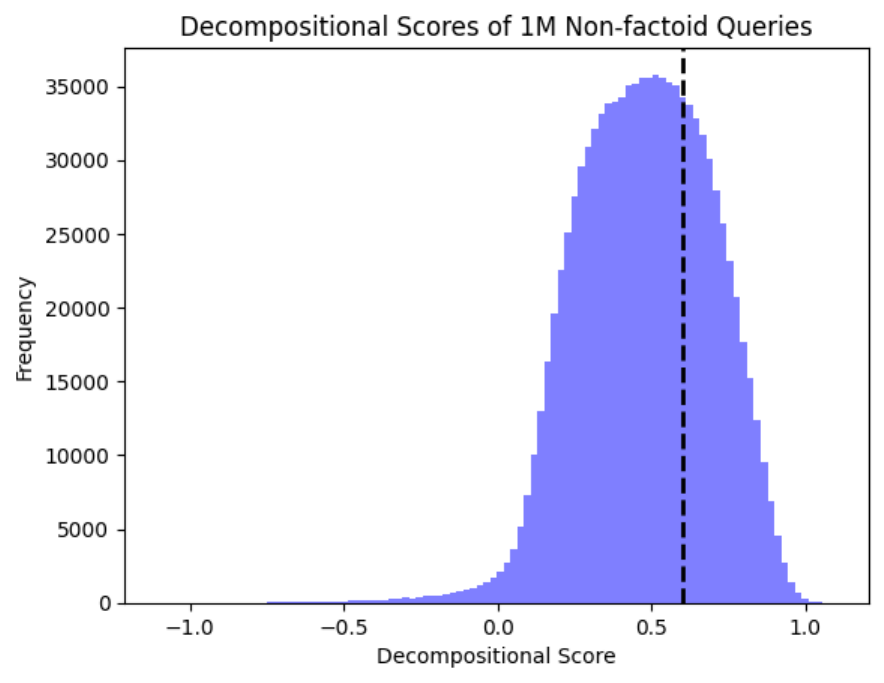}
  \label{fig:sub2}
\end{subfigure}
\caption{(Left) Non-factoid scores of the 15.7M QnA Queries. The roughly 1M queries whose score exceeded the threshold +0.75 were sent to the Decompositional classifier. Note that because this was a binary classifier, 89\% of the non-factoid scores were less than -0.75, which is cut off from the left-hand histogram to make it easier to visualize. (Right) The Decompositional classifier's scores of the roughly 1M Non-factoid queries. Around 146k queries exceeding the 0.6 threshold line resulted are considered both Non-factoid and Decompositional, and were then de-duplicated to arrive at the final \dataset dataset of around 100k.}
\label{fig:histograms_factoid_decomposition}
\end{figure*}

\begin{table}[ht]
\centering
\small
\begin{tabular}{|p{10cm}|c|c|}
\hline
\textbf{Question} & \textbf{Factoid} & \textbf{Decompositional ($\downarrow$)} \\
\hline
Why is it illegal to destroy money & 1.02 & 0.59 \\ \hline
Which law firms offer the best opportunities for international work? & 1.07 & 0.58 \\ \hline
How should a grad student communicate with faculty & 0.80 & 0.56 \\ \hline
How does a short story differ from a novel & 1.07 & 0.54 \\ \hline
Why is Beowulf an important literary work & 1.10 & 0.51 \\ \hline
Why are e scooters illegal & 0.90 & 0.50 \\ \hline
Why did the declaration of independence start & 1.12 & 0.49 \\ \hline
How did marsupials evolve & 1.02 & 0.48 \\ \hline
Is turbulence dangerous & 0.97 & 0.47 \\ \hline
Why do people cut down the Amazon rainforest & 1.09 & 0.45 \\ \hline
Why are flags at half mast in Indiana today & 1.06 & 0.45 \\ \hline
Why the houses in Rajasthan have thick walls and flat roofs? & 1.06 & 0.43 \\ \hline
How do forensic pathologists determine cause and manner of death? & 1.02 & 0.43 \\ \hline
What causes changes in the moon's appearance & 1.10 & 0.41 \\ \hline
How did Edgar Allan Poe start his writing career & 0.97 & 0.40 \\ \hline
How common was vasectomy in the 1950's & 0.90 & 0.40 \\ \hline
Why are behavioral questions important & 0.79 & 0.38 \\ \hline
Is pepper spray harmful & 0.93 & 0.37 \\ \hline
How do fertilizers increase productivity & 1.05 & 0.35 \\ \hline
How do we get matter and energy from food & 1.11 & 0.32 \\ \hline
What type of competition occurs when only a few firms dominate a market? & 1.08 & 0.31 \\ \hline
Why can't waves travel through a vacuum? & 1.08 & 0.30 \\ \hline
What caused the Dixie fire & 0.84 & 0.29 \\ \hline
Why camels live in desert & 1.02 & 0.29 \\ \hline
How was COVID-19 discovered & 0.92 & 0.28 \\ \hline
How are elements organized into groups & 0.96 & 0.27 \\ \hline
Why did Montezuma give the Spanish gold? & 1.01 & 0.27 \\ \hline
Differences between debt and equity financing & 0.98 & 0.27 \\ \hline
How are fossils older than 60,000 years dated? & 1.08 & 0.17 \\ \hline
What happens when a data collision occurs on an Ethernet bus? & 1.03 & 0.16 \\ \hline
How is gold formed in the earth's crust & 0.98 & 0.12 \\
\hline
\end{tabular}
\caption{\label{tab:nonfactoid_but_not_quite_decompositional}
Examples of QnA Queries that were non-factoid (had a non-factoid classifier score above 0.75), but were \textbf{not} decompositional (had decompositional classifier score \textbf{less} than 0.6). Further down this list, more of the examples have a single correct answer (even if it is somewhat long-form), but there is obviously a lot of gray area,  highlighting the challenges of trying to filter large amounts of queries by simple thresholds. }
\end{table}

\begin{figure*}
\begin{framed}
\#\#\# For the following (Question | score | reason) triples, the score indicates how "good" of a non-factoid question they are in the sense that they can lead to interesting and in-depth analysis. \\
\#\#\# \textbf{Definition}: A good non-factoid question is specific, with potential to amount to a good research report with a clear and refutable thesis, supported by evidence and analysis.  \\
\#\#\# \textbf{Characteristic formats} of good non-factoid questions (not exhaustive):
\begin{itemize}[nolistsep]
    \item Good non-factoid questions will often talk about the relationship between two things, e.g. "Compare and contrast X and Y", "How/why does X affect/impact Y?", "Why X is significant to Y", or "What role does X play in Y?", or "to what extent does X lead to Y?", etc.
    \item A good non-factoid question can also ask "Why does X happen", "What factors play a role in X?", "How is X significant" or "What is the cause of X", but it should be specific about what kind of analysis is expected.
    \item Other forms of good non-factoid questions can ask about the pros/cons, benefits/detriments of something, or compare/contrast two things, etc.
\end{itemize} 
\#\#\# \textbf{Instructions}: Rate each question on a scale of 0-10, where 0 is a factoid question and 10 is an excellent non-factoid question and then provide a brief reason for your rating \\
Q: how tall is abraham lincoln | 0 | factoid \\
Q: can i change the weather | 2 | personal question \\
Q: was the civil war fought over slavery | 5 | fair, but could more directly ask about other important facets of the causes of the civil war and their role in the conflict \\
Q: to what extent was the civil war fought over slavery | 8 | good, will lead to in-depth analysis on the causes of the civil war \\
Q: what impact do human activities have on the weather | 10 | excellent, many in-depth reports written to answer this question \\
Q: should LA invest more in railway or highway infrastructure for public transport | 9 | great \\
Q: what is an example of blackbody radiation? | 0 | asking for an example \\
Q: could not determine type for | 0 | not a question \\
Q: what typically signals the end of the olympic games | 2 | factoid, olympic closing ceremony can be looked up easily \\
Q: Why were Navajo code talkers used during WW2? | 7 | good, could lead to analysis of how culture and language can be used in warfare \\
Q: When does protein folding begin? | 1 | has a single, known correct answer \\
Q: what is the cost and necessary materials to build a refinery | 5 | fair, asks about a complex process but will not likely elicit analysis \\
Q: What is the Navavidha Bhakti? | 0 | asking for a definition \\
Q: why is technological change bad? | 5 | fair, but could be more specific \\
Q: analyze how technological changes have historically impacted cultures | 10 | excellent, very specific \\
Q: who owns phone number 280-626-1435 | 0 | personally identifiable information \\
Q: What are the main differences between regulations of the NFL and the CFL? | 4 | has potential for in-depth analysis but doesn't explicitly ask for it \\
Q: Why do planes using rivets \& not welded construction? | 7 | good, will require in-depth analysis on aerospace technology \\
Q: How did the Catholic Pope manage to become more powerful than Kings in old Europe? | 9 | much potential for historical analysis \\
Q: interesting facts about korea | 0 | not specific \\
Q: \{Question\} |  

\end{framed}
\caption{Prompt given to text-davinci-003 to collect labels of whether a question is non-factoid. The current question is substituted at the end. Labels on 1-10 were binarized based to train the non-factoid classifier.}
\label{fig:promp-label-nonfactoid}
\end{figure*}

\begin{figure*}
\begin{framed}
Question: \{Question\} \\
\#\#\# Instructions: how clearly does the above question require sub-questions in order to answer it? Think silently about how complex or multi-facted the question is. Imagine you are a person trying to answer the question using a search engine like Google. Would the person likely have to issue multiple queries to arrive at a comprehensive answer? Would they need to exert more effort to fully understand the nuances behind the question, or different perspectives surrounding it? etc. Or, is the person likely to find the best answer with a single search with low effort? Give your rating on 1-100 between <score> and </score> tags, where 1 means the question is trivial or common knowledge, while 100 means it likely involves effort to break the question down into many sub-questions or facets.
<score>

\end{framed}
\caption{Prompt given to gpt-35-turbo to collect labels of how appropriate a question is for ``decomposition'' into sub-questions. These labels were used to train the Decompositional classifier. }
\label{fig:promp-label-decompositional}
\end{figure*}

\label{sec:appendix}

\begin{figure*}
    \centering
    \includegraphics[width=\linewidth]{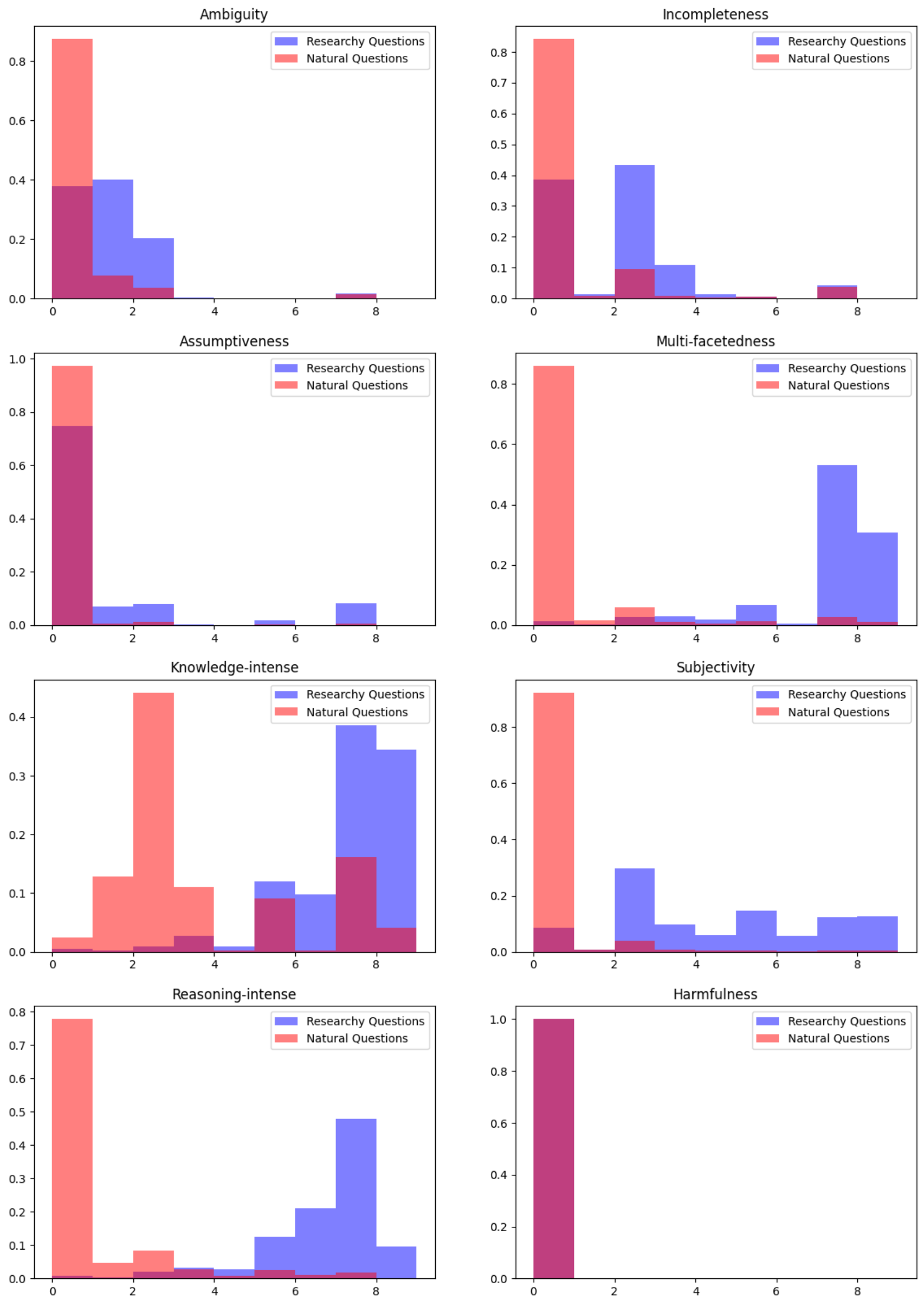}
    \caption{We had GPT-4 label all 102k Researchy question -- and a uniform sample of 2k Natural Questions -- along 8 dimensions using the prompt in Figure~\ref{fig:gpt4-classifier-prompt}. Each dimension was on a scale of 1-10; we plot their normalized densities. For Researchy Questions, these are the scores of the final 96k released questions after final filtering described in Section~\ref{sec:stage-5-filtering}. Lower scores are better for Ambiguity, Incompleteness, Assumptiveness and Harmfulness. We intended Researchy Questions to have higher scores for Knowledge-intensiveness, Reasoning-intensiveness and Multi-facetedness.}
    \label{fig:autolabeling_histograms_gpt}
\end{figure*}

We had GPT-4 label certain attributes of each Researchy Question using the prompt given in Figure~\ref{fig:gpt4-classifier-prompt}. Particularly, we focus on removing questions that were too \textbf{incomplete} or underspecified to meaningfully answer (shown in Table~\ref{tab:example_incomplete_queries}), or those that are too \textbf{presumptuous} (shown in Table ~\ref{tab:example_assumptive_queries}). Incomplete queries could arise when users ask queries that e.g. refer to previous queries or subjects in their session. 

\begin{table*}[bh]
\begin{tabular}{|p{0.475\linewidth}|p{0.475\linewidth}|} 
\hline 
\textcolor{red}{how can climate change} & \textcolor{red}{why did the market crash} \\ \hline
\textcolor{red}{are perspectives that change when presented with fresh experiences} & \textcolor{red}{how did the war for independence end
} \\ \hline
\textcolor{red}{in what ways did native peoples transform} & \textcolor{red}{how dangeous can a persona affect a person personality} \\ \hline
\textcolor{red}{how would the law treat the issues confronting a diverse family?} & \textcolor{red}{how did president lyndon b. johnson respond to this event?} \\ \hline
\textcolor{red}{compare and contrast social contract} & \textcolor{red}{what is the difference in how the government reacted in these two situations?} \\ \hline
\textcolor{red}{how were others influenced by miyamoto} & \textcolor{red}{explain why summers were difficult for europeans.} \\ \hline
\textcolor{red}{what effect did the recent looting have on emigration} & \textcolor{red}{what is the main reason why some teams never reach stage 4} \\
\bottomrule
\end{tabular}
\caption{\label{tab:example_incomplete_queries} Examples of incomplete or ambiguous queries a.k.a too underspecified to attempt to answer meaningfully (roughly 3\% of queries in the last stage of filtering) that were \textbf{removed} from the dataset by additional GPT-4 filtering.} 
\end{table*}

\begin{table*}[bh]
\begin{tabular}{|p{0.475\linewidth}|p{0.475\linewidth}|} 
\hline 
\textcolor{red}{how college became a ruthless competition divorced from learning} &
\textcolor{red}{why is airline customer service so bad} \\ \hline
\textcolor{red}{why are police departments afraid of change} & \textcolor{red}{why immigration is bad for the us} \\ \hline
\textcolor{red}{why did chinese families not desire to have female children?} & \textcolor{red}{why do games make people socially disconnected from people} \\ \hline
\textcolor{red}{why are catholics democrats} & \textcolor{red}{why junior doctors are treated so badly} \\ \hline
\textcolor{red}{how casinos have ruined towns} & \textcolor{red}{why should zoos be shut down} \\ \hline 
\textcolor{red}{how is the government controlling us} & \textcolor{red}{why rehabilitation does not work in america} \\ \hline
\textcolor{red}{why veganism is bad for animals} & \textcolor{red}{why most clinical research is not useful} \\
\bottomrule
\end{tabular}
\caption{\label{tab:example_assumptive_queries} Examples of ``presumptious'' queries we removed from the dataset (roughly 2\% of queries in the last stage of filtering), roughly defined as those where the question is slanted or biased in some way. Note, not all assumptive queries were harmful.} 
\end{table*}

\begin{figure*}
\begin{framed}
Given the question: \{Question\} \\
\\
Instructions: Please output a python dictionary with fields scoring the question on the following criteria: \\
\begin{enumerate}[nolistsep]
    \item "\textbf{ambiguous}" : Int 0-10  to what extent is the intent of the question ambiguous (has more than one interpretation); 0 means no major ambiguity. Not to be confused with subjectiveness or incompleteness.
    \item "\textbf{incompleteness}" : Int 0-10  indicating how difficult it is to determine the intent of the question, whether it is missing crucial context or details that ought to be specified in order to answer the question; 0 means the question is answerable and self-contained, 10 means the question is un-answerable because it is incomplete or under-specified.
    \item "\textbf{assumptive}" : Int 0-10 the degree to which the question has built-in assumptions or biases (that are not offensive, which is point 8 below); 0 means no notable or unreasonable assumptions.
    \item "\textbf{multi-faceted}" : Int 0-10 the degree to which the question has multiple facets or perspectives that need to be considered in order to answer it; 0 means the question is straightforward and has a single, undisputed answer.
    \item "\textbf{knowledge-intensive}" : Int 0-10 the degree to which the question would require specialized knowledge (like textbooks, scholarly articles, etc) to provide a thorough and grounded answer; 0 means the answer is common knowledge or can be looked up instantly in common references, 10 means the questions probably entails a lot of work to find and analyze specialized knowledge. 
    \item "\textbf{subjective}" : Int 0-10 the degree to which the question is subjective, meaning an answer(s) exist, but there is no agreed-upon way to determine which one is better; 0 means the question is largely objective i.e. the overwhelming majority of people would agree on the answer if they knew it.
    \item "\textbf{reasoning-intensive}" : Int 0-10 the degree to which the question requires reasoning to synthesize an answer; 0 means the question can be answered trivially e.g. by looking up a fact, referencing an encyclopedia or database, or using a calculator (once).
    \item "\textbf{harmful}" : Int 0-10 to what extent the question could be interpreted as being harmful (physically or psychologically to oneself, others, or animals), offensive, overly biased, sexually explicit, or otherwise inappropriate for e.g. someone of the age of 12 to be exposed to. \\
\end{enumerate}
Note that the above criteria are not mutually exclusive, e.g. a question can be both subjective and knowledge-intensive, for example "is capitalism better than socialism" would be both.
Make sure to output only the valid python dictionary without comments or other extraneous output.

\end{framed}
\caption{Prompt given to GPT-4 to classify attributes of each of the Researchy Questions. The output labels were used to further filter the data for release.}
\label{fig:gpt4-classifier-prompt}
\end{figure*}

\subsection{Additional Safety Filtering}
In addition to the official offensiveness/adult query classifiers used in our commericial search engine, we also had GPT-4 label on a scale of 1-10 ``\textit{to what extent the question could be interpreted as being harmful (physically or psychologically to oneself, others, or animals), offensive, overly biased, sexually explicit, or otherwise inappropriate for e.g. someone of the age of 12 to be exposed to.}''. We found roughly 2\% of queries had an unsatisfactory harmfulness score and hence were removed from the released dataset.  

Table~\ref{tab:example_harmful_queries} shows some examples of those removed. Real people asked these questions: we acknowledge that some of these may be motivated by genuine information-seeking intent, notwithstanding, we believe there is too high a risk that the act of answering these questions may lead to harm.

\begin{table*}[bh]
\begin{tabular}{|p{0.475\linewidth}|p{0.475\linewidth}|} 
\hline 
\textcolor{red}{Is it fair to blame younger people for covid spike} & \textcolor{red}{How bad is illegal immigration} \\ \hline
\textcolor{red}{Why do people pretend to be depressed} & \textcolor{red}{Why veganism is bad for animals} \\ \hline
\textcolor{red}{what factors affect when deciding to terminate a down sydrome child} & \textcolor{red}{Why is the business of sex so popular and lucrative?} \\ \hline
\textcolor{red}{how people make money illegally} & \textcolor{red}{Was the holocaust real?} \\ \hline
\textcolor{red}{Why was gender queer banned} & \textcolor{red}{Which countries persecute christians} \\ \hline
\textcolor{red}{Why do so many women get abortions} & \textcolor{red}{Was the holocaust worse than slavery} \\ \hline
\textcolor{red}{Could a small nuclear war reverse global warming} & \textcolor{red}{How do holocaust deniers try to gain followers?} \\ \hline
\textcolor{red}{Is china a threat to world peace} & \textcolor{red}{Why christians believe conspiracy theories} \\ \hline
\textcolor{red}{Why do so many incompetent men become leaders} & \textcolor{red}{Why did africans sell other africans into slavery} \\ \hline
\textcolor{red}{Why is bribery being used to force people to take vaccine} & \textcolor{red}{Why were many whites scared away from supporting civil rights} \\ \hline
\textcolor{red}{What makes a church a cult} & \textcolor{red}{Why did poor southerners fight for slavery} \\ \hline
\textcolor{red}{Are minorities more likely to commit crime} & \textcolor{red}{Why isn't christianity considered a cult} \\ \hline
\textcolor{red}{Why corporal punishment should be legal} & \textcolor{red}{Pros and cons of sexism} \\ \hline
\textcolor{red}{Should we stratify people based on intelligence or other specific qualities?} & \textcolor{red}{Why are we challenged by the arrival of migrants in our country?} \\ \hline
\textcolor{red}{Why are we protecting the unvaccinated} & \textcolor{red}{Is the death penalty racially motivated} \\ \hline
\textcolor{red}{Why is bayonetta so sexualized} & \textcolor{red}{Why is racial profiling important in society} \\ \hline
\textcolor{red}{Does islam promote violence} & \textcolor{red}{Is affirmative action racism} \\ \hline
\textcolor{red}{Who is behind mass immigration} & \textcolor{red}{Why is the vaccine killing people} \\ \hline
\textcolor{red}{Why is the west so intent in bringing in so many illegal migrants} & \textcolor{red}{Why might the artist have thought reconstruction was “worse than slavery” for black southerners?} \\  
\bottomrule
\end{tabular}
\caption{\label{tab:example_harmful_queries} Here we show some examples of the roughly 2\% of queries that were \textbf{removed} from the dataset by additional GPT-4 harmfulness filtering.} 
\end{table*}

\begin{figure*}
\small
\begin{framed}
Question: \{Question\}
\\

\#\#\#\#\#\#\#\#\#\#\#\#\#\#\#\#\#\#\#\#\# \\
Reference Answer (assumed to be true): \{ReferenceAnswer\} \\

\{CandidateAnswers\}
\\

\#\#\#\#\#\#\#\#\#\#\#\#\#\#\#\#\#\#\#\#\# \\
Keep in mind the following Guidelines when evaluating the Candidate Answers: \\

\textbf{Guidelines}: 
\\
\begin{itemize}[nolistsep]
    \item \textbf{Usefulness}: A useful answer adds value by providing in-depth analysis, actionable steps, and relevant information that helps users make informed decisions, complete tasks, and understand complex problems in a simpler way. It also considers real-world scenarios, cause and effect relationships, and historical context to enhance the user's understanding.
    \item \textbf{Relevance}: A relevant answer directly addresses the question's intent in a sensible and specific manner, without being overly broad, redundant, or distracting. It should be clear, on-topic, and provide helpful information tailored to the question asked.
    \item \textbf{Truthfulness} in an answer means providing accurate and verifiable information, without making false claims, unverifiable statements, or promoting misleading information. It should agree with the Reference Answer. Contradicting the Reference Answer *must* result in lower score.
    \item \textbf{Groundedness} in an answer means providing information supported by reliable sources and following a logical chain of reasoning, instilling confidence in its accuracy. The answer should be based on credible evidence, address uncertainties, and avoid relying on personal opinions, rumors, or hearsay.
    \item \textbf{Thoroughness} involves considering the full scope of a question, addressing multiple perspectives, scenarios, and outcomes, and ensuring all important aspects are covered without omission. It requires analyzing pros and cons, empathizing with various stakeholders, and addressing different interpretations or facets of a question. \\
\end{itemize}

\#\#\#\#\#\#\#\#\#\#\#\#\#\#\#\#\#\#\#\#\#
\textbf{Instructions}: Above are \{NumCandidates\} Candidate Answers to the Question: \{Question\}. Please read them carefully along with the Guidelines for how to evaluate an answer's quality. Then: 
\begin{enumerate}[nolistsep]
    \item Briefly \textbf{critique} each Candidate Answer with respect to the Guidelines, formatted in *bullet points* between "<CritiqueX>" and "</CritiqueX>" tags where X is the Candidate Answer number. Each bullet is between sub-tags of either <strength>, <weakness>, or <missinginfo>. A <strength> is where the Answer makes a good point that follows the Guidelines and contributes to a helpful response. A <weakness> is where the Answer makes an incorrect, irrelevant, unreasonable or overly broad claim that fails to address the full scope of the Question, or otherwise falls short of the Guidelines. <missinginfo> is when some key event, facts or other material information is omitted but should have included to strengthen the Candidate Answer.
    \item  Give a \textbf{binary correctness} score indicating whether each Candidate Answer generally agrees with the Reference Answer -- "<CorrectX>1</CorrectX>" -- or contradicts it "<CorrectX>0</CorrectX>", where X is the Candidate Answer number.
    \item \textbf{Explain} which Candidate Answer is better and why, i.e. how well it adheres to the Guidelines, between "<Explanation>" and "</Explanation>" tags."
    \item Finally, \textbf{score} each Candidate Answer on 1-100, where 100 is a perfect Answer that aligns with the Guidelines. Indicate the score between "<ScoreX>" and "</ScoreX>" tags where X is the Candidate Answer number.
\end{enumerate}

Make sure to address all \{NumCandidates\} Candidate Answers. When you are finished, conclude your response with "=====".

<Critique1>

\end{framed}
\caption{Prompt given to GPT-4 to act as a ``critique-then-score'' judge of candidate answers side-by-side. }
\label{fig:critique-then-score-prompt}
\end{figure*}

\begin{table*}[bh]
\small
\renewcommand{\arraystretch}{1.5} 
\begin{tabular}{|c|}
\hline
\multicolumn{1}{|l|}{\textbf{Researchy Question}: how does public transport help the economy}  \\ \hline
\multicolumn{1}{|c|}{\textbf{GPT-4 Turbo Closed-book Answer} (2023-07-01-preview)} \\ \hline
\begin{minipage}[t]{\columnwidth}
{Public transport provides numerous economic benefits to communities and countries as a whole. Here are some of the key ways in which public transportation supports economic growth and efficiency: \\ 
\begin{enumerate}[nolistsep]
    \item \textbf{Reduced Congestion Costs:} Public transport can significantly reduce the number of vehicles on the road, which in turn reduces traffic congestion. This leads to a more efficient transportation system and increased productivity.  
    \item \textbf{Cost Savings for Individuals:} Public transport can be more cost-effective than owning a private vehicle, leading to increased disposable income for individuals.  
    \item \textbf{Job Creation:} Public transport systems create jobs for operation, maintenance, and construction, contributing to employment opportunities.  
    \item \textbf{Increased Property Values:} Good public transport connectivity often increases property values, leading to higher tax revenues.  
    \item \textbf{Environmental Benefits:} Public transport reduces the number of cars on the road, decreasing air pollution and greenhouse gas emissions.  
    \item \textbf{Tourism:} Efficient public transport systems enhance tourism by making it easier for tourists to navigate and spend in local economies.  
    \item \textbf{Business Productivity:} Reliable public transport reduces employee tardiness and absenteeism, increasing business productivity.  
    \item \textbf{Social Inclusion:} Public transport provides mobility for those without access to private vehicles, allowing for greater workforce participation and access to services.  
    \item \textbf{Urban Development:} Public transport encourages efficient land use and can be a catalyst for urban revitalization and development.  
    \item \textbf{Energy Efficiency:} Public transport is more energy-efficient per passenger kilometer than private vehicles, contributing to energy security.  
\end{enumerate}  
By providing an affordable and efficient alternative to private vehicles, public transport systems can enhance the overall economic performance of a region and improve the quality of life for its residents.
} 
\end{minipage} \\ \hline 
\multicolumn{1}{|c|}{\textbf{Mixtral 8x7b Closed-book Answer} (\url{https://stablediffusion.fr/mixtral})} \\ \hline
\begin{minipage}[t]{\columnwidth}
Public transportation can have a positive impact on the economy in several ways:
\begin{enumerate}[nolistsep]
    \item Reduced traffic congestion: Public transportation helps to reduce traffic congestion by providing an alternative to driving, which can lead to decreased travel times and increased productivity. This can also reduce the need for costly road expansion projects.
    \item Increased economic activity: Public transportation can help to stimulate economic activity by making it easier for people to access jobs, shops, and other services. This can lead to increased spending and investment in local communities.
    \item Lower transportation costs: Public transportation can provide a more affordable transportation option for many people, especially those who cannot afford to own and maintain a car. This can help to reduce household transportation costs and free up money for other expenses.
    \item Improved air quality: Public transportation can help to reduce air pollution by taking cars off the road. This can lead to improved public health and reduced healthcare costs.
    \item Increased property values: Properties located near public transportation hubs, such as train stations and bus stops, often have higher property values than those located further away. This can lead to increased tax revenues for local governments.
    \item Job creation: Public transportation agencies and companies create jobs for drivers, mechanics, engineers, and other professionals. These jobs can provide stable employment and contribute to the local economy.
    \item Energy efficiency: Public transportation is generally more energy-efficient than individual car travel. This can help to reduce dependence on fossil fuels and decrease greenhouse gas emissions.
\end{enumerate}
Overall, public transportation can play an important role in supporting a strong and sustainable economy by providing affordable, efficient, and environmentally-friendly transportation options. \\ 
\end{minipage} \\ \hline
\end{tabular}

\caption{\label{tab:closed-book-examples} Closed-book direct answers from GPT-4 and Mixtra 8x7b~\cite{jiang2024mixtral} to the Researchy Question used in Table~\ref{tab:example_decomposition}. We find these two answers to be strikingly similar (suggesting convergent evolution of LLMs), and both paint in very broad strokes an answer that isn't ``wrong'' but rather, superficial.} 
\end{table*}

\begin{figure*}
\small
\begin{framed}
Public transportation plays a significant role in boosting the economy in several ways:

\begin{enumerate}[nolistsep]
\item Job Creation: Investment in public transportation can create millions of jobs. For instance, every \$1 billion invested could create 50,000 jobs [1,2].
\item Economic Returns: Every \$1 invested in public transportation could generate \$5 in economic returns [1].
\item Reduced Traffic Congestion: A well-functioning public transit system reduces traffic congestion, leading to more efficient travel and productivity [3].
\item Economic Growth: Public transportation can shape land use and development patterns, stimulate commerce, and increase property values [4].
\item Supports Vulnerable Populations: It improves equality, particularly for the more vulnerable members of society, by providing them with mobility and access to jobs [1].
\item Environmental Impact: Public transit helps improve air quality and reduce greenhouse gas emissions, contributing to a sustainable future [3].
\end{enumerate}

In summary, public transportation is more than just a way to move people around. It’s a vehicle for opportunity, equity, and a better quality of life.

\textbf{References} 
\begin{itemize}[nolistsep]
    \item \href{https://www.weforum.org/agenda/2021/04/here-s-why-cities-should-invest-in-public-transport/}{[1] https://www.weforum.org/agenda/2021/04/here-s-why-cities-should-invest-in-public-transport/}   
    \item \href{https://infrastructureusa.org/the-economic-impact-of-public-transportation/}{[2] https://infrastructureusa.org/the-economic-impact-of-public-transportation/}
    \item \href{https://www.ncesc.com/is-public-transit-better-for-the-economy/}{[3] https://www.ncesc.com/is-public-transit-better-for-the-economy/}
    \item \href{https://www.apta.com/wp-content/uploads/Resources/resources/reportsandpublications/Documents/Economic-Recovery-APTA-White-Paper.pdf}{[4] https://www.apta.com/wp-content/uploads/Resources/resources/reportsandpublications/Documents/Economic-Recovery-APTA-White-Paper.pdf}
    \item \href{https://profound-tips.com/advice/how-does-public-transportation-help-the-economy/}{[5] https://profound-tips.com/advice/how-does-public-transportation-help-the-economy/}
\end{itemize} 

\end{framed}
\caption{A ``screenshot'' of the output from Bing Chat in precise mode (a.k.a. Microsoft Copilot) for the question ``how does transportation help the economy''. }
\label{fig:bingchat_example}
\end{figure*}

\begin{figure*}
\small
\begin{framed}
\textbf{Comprehensive Research Report: The Role of Transportation in Economic Development} \\
\textbf{\# Introduction}
Transportation is a critical factor in the economic development of any region or nation. It serves as the backbone of economic growth, facilitating the movement of goods, services, and labor, which are essential for the efficient functioning of the economy. This report synthesizes key points from various sources to illustrate how transportation contributes to economic prosperity. \\

\textbf{\# Key Contributions of Transportation to the Economy
}
\textbf{\#\# Job Creation and Economic Growth
}
Investments in transportation infrastructure, such as public transit, are directly linked to job creation and economic growth. For every \$1 billion invested in transit, an estimated 49,700 to 50,731 jobs can be created, offering a significant return on investment ranging from 4 to 1 to as high as 5 to 1 . This investment not only supports American jobs and industry but also generates income for local businesses and their workers . Moreover, spending on transportation programs creates jobs in the short term, particularly in construction, and can stimulate the economy . \\

\textbf{\#\# Enhancing Business Productivity and Competitiveness
}
Transportation improvements lead to reduced travel times and costs, as well as increased network reliability, which in turn enhances business productivity and competitiveness . By reducing transport costs, businesses can access higher-quality or lower-cost inputs and connect with potential suppliers . This connectivity also allows individuals to access a wider range of jobs, thereby participating more fully in the labor market . \\

\textbf{\#\# Market Accessibility and Economic Activity
}
An efficient transport system provides market accessibility by linking producers and consumers, enabling transactions to occur . This linkage is crucial for the economy as transportation accounts for a significant portion of the GDP, ranging from 6\% to 12\% . By offering cost, time, and reliability advantages, transportation enables goods to be transported over longer distances, fostering economic changes, most of which are positive . \\

\textbf{\#\# Induced Investment and Regeneration Impacts} 
Transportation investments can influence the level or location of private sector investment due to improved connectivity . These investments can also lead to local economic impacts resulting from an improved local image and the attraction of land use development, known as regeneration impacts . \\

\textbf{\#\# Structural and Developmental Transformations}
Sustained improvements in transportation, along with parallel advancements in information and production technologies and institutional structures, can lead to structural and developmental transformations within the economy . This demonstrates the dynamic nature of transportation's role in economic development. \\

\textbf{\#\# Economic Efficiency and Scale}
An efficient transportation system reduces costs across many economic sectors, while an inefficient one increases these costs . Moreover, more efficient transport costs less, leading to rising productivity and economies of scale, which should cause GDP to rise faster than transport activity . \\

\textbf{\#\# Influence on Economic Development and Military Strength}
Transportation activities contribute to a nation's total economic product and strengthen the national or regional economy . A well-developed transportation network not only makes markets more competitive but also adds to a nation's military strength . \\

\textbf{Conclusion}
Transportation plays a pivotal role in the economic development of a country. It is essential for job creation, enhancing business productivity, providing market accessibility, inducing investment, and contributing to structural transformations. Governments recognize the importance of investing in transportation infrastructure as a means to stimulate economic growth and maintain a competitive edge in the global market. As such, transportation is not just a means of moving people and goods; it is a fundamental driver of economic prosperity. \\

\textbf{References} 
\begin{itemize}[nolistsep]
    \item \href{https://www.apta.com/research-technical-resources/economic-impact-of-public-transit/}{Economic Impact of Public Transit. (2019, April 22). American Public Transportation Association.}  
    \item \href{https://greener-vision.com/guidance-tool/relationship-between-transport-economy/ }{The relationship between transport and the economy. (2017, February 13). Greener Vision.} 
    \item \href{https://www.apta.com/research-technical-resources/research-reports/economic-impact-of-public-transportation-investment/}{Economic Impact Of Public Transportation Investment. (2019, March 27). American Public Transportation Association.}
    \item \href{https://transportgeography.org/contents/chapter3/transportation-and-economic-development/ }{3.1 – Transportation and Economic Development | The Geography of Transport Systems. (2017, December 4). Transport Geography.}
    \item \href{https://www.accessmagazine.org/spring-2005/economic-consequences-transport-improvements/}{Economic Consequences of Transport Improvements. (2022, October 11). ACCESS Magazine.}
    \item \href{https://www.accessmagazine.org/spring-2011/transportation-jobs-economic-growth/}{Transportation, Jobs, and Economic Growth. (2018, February 14). ACCESS Magazine. }
    \item \href{https://www.britannica.com/money/topic/transportation-economics }{Transportation economics | Definition, Importance, \& Facts. (1998, September 8). Britannica. }
    \item \href{https://www.futurelearn.com/info/courses/what-is-economics-in-the-world-of-global-logistics/0/steps/95653 }{The role of transport in economic development. FutureLearn.} 
\end{itemize}

\end{framed}
\caption{A ``screenshot'' of the output from YouPro Research engine for the query ``how does transportation help the economy''. \url{https://you.com/search?q=how+does+transportation+help+the+economy}}
\label{fig:youpro_research_example}
\end{figure*}

\begin{figure*}
\small
\begin{framed}
\textbf{Query \#1 - how does transportation help the economy} \\
\textbf{Query \#2 - How does transportation infrastructure affect economic growth?} \\
\textbf{Query \#3 - What are the long-term effects of transportation investments?} \\
\textbf{Query \#4 - How does transportation contribute to GDP?} \\
\textbf{Query \#5 - What are the environmental impacts of transportation?} \\
\textbf{Query \#6 - How does transportation affect urban planning and development?} \\
\textbf{Query \#7 - What are the challenges in financing transportation infrastructure?} \\
\textbf{Query \#8 - How can transportation policy influence economic productivity?} \\

\textbf{Final Research Result for how does transportation help the economy} \\
Transportation plays a pivotal role in the economy, contributing to the movement of goods, services, labor, and capital, which are essential for the functioning of a modern society and the growth of an economy [11]. The ability to transport goods and people efficiently is fundamental to economic life, as seen in the early United States where a robust transportation network was instrumental in the country's development. In a macroeconomic sense, transportation activities are a part of the total economic product, and the demand for transportation services contributes to the GDP [2]. The transportation sector not only supports the movement of goods and services but also affects employment, wages, and business income, thereby contributing to the overall economic growth [3].   \\

Investments in transportation infrastructure can lead to job creation and improved connectivity, which in turn, promotes economic productivity and well-being [22]. Furthermore, efficient transportation systems can reduce poverty by increasing access to employment, recreational, and other essential life-shaping opportunities [20]. The economic importance of transportation is also evident in the fact that transportation-related final demand has been increasing since the recession, indicating a strong correlation between transportation and economic activity [8]. Additionally, the World Bank and ITDP report highlights the importance of active mobility in promoting economic productivity and socially inclusive development [21].   \\

Transportation is a critical component of a society, reflecting the aspirations of a society and influencing the location decisions of people and businesses. It is a vital determinant of economic growth, job creation, and improved living conditions, and its efficiency can lead to a green future by reducing congestion and promoting sustainable development [19]. \\  

In conclusion, transportation is not only crucial for the movement of goods and services but also plays a significant role in the overall economic growth of a country. It contributes to the GDP, supports employment, and affects wages and business income [2][3]. Investments in transportation infrastructure can lead to job creation and improved connectivity, promoting economic productivity and well-being [22]. Additionally, transportation plays a vital role in reducing poverty by increasing access to essential opportunities [20]. The economic importance of transportation is evident in the strong correlation between transportation and economic activity [8]. Therefore, transportation is a critical component of a society, reflecting its aspirations and influencing the location decisions of people and businesses [19]. \\

\textbf{References} 
\begin{itemize}[nolistsep]
    \item \href{https://www.britannica.com/money/topic/transportation-economics}{[2] https://www.britannica.com/money/topic/transportation-economics}   
    \item \href{https://data.bts.gov/stories/s/Transportation-Economic-Trends-Contribution-of-Tra/smrm-36nv/}{[3] https://data.bts.gov/stories/s/Transportation-Economic-Trends-Contribution-of-Tra/smrm-36nv/}
    \item \href{https://www.bts.gov/sites/bts.dot.gov/files/docs/browse-statistical-products-and-data/transportation-economic-trends/215926/transportation-economic-trends-2017-chapter-2.pdf}{[8] https://www.bts.gov/sites/bts.dot.gov/files/docs/browse-statistical-products-and-data/transportation-economic-trends/215926/transportation-economic-trends-2017-chapter-2.pdf }
    \item \href{https://www.nationalgeographic.org/article/effects-transportation-economy/5th-grade/}{[11] https://www.nationalgeographic.org/article/effects-transportation-economy/5th-grade/}
    \item \href{https://transportgeography.org/contents/chapter3/transportation-and-society/}{[19] https://transportgeography.org/contents/chapter3/transportation-and-society/}
    \item \href{https://blogs.iadb.org/transporte/en/how-can-transport-reduce-poverty-and-promote-socially-inclusive-development/}{[20] https://blogs.iadb.org/transporte/en/how-can-transport-reduce-poverty-and-promote-socially-inclusive-development/}
    \item \href{https://www.usatoday.com/story/special/contributor-content/2024/01/10/itdp-and-world-bank-report-highlights-economic-and-environmental-benefits-of-financing-active-mobili/72177281007/}{[21] https://www.usatoday.com/story/special/contributor-content/2024/01/10/itdp-and-world-bank-report-highlights-economic-and-environmental-benefits-of-financing-active-mobili/72177281007/}
    \item \href{https://www.accessmagazine.org/spring-2011/transportation-jobs-economic-growth/}{[22] https://www.accessmagazine.org/spring-2011/transportation-jobs-economic-growth/}
\end{itemize} 

\end{framed}
\caption{A ``screenshot'' of the output from SciPhi Research engine for the query ``how does transportation help the economy''. This research engine also asked and separately answered multiple queries before arriving at the final research result. \url{https://search.sciphi.ai/research?q=how+does+transportation+help+the+economy}}
\label{fig:sciphi_research_example}
\end{figure*}

\end{document}